\documentclass[letterpaper, 10 pt, conference]{ieeeconf}
\IEEEoverridecommandlockouts
\let\labelindent\relax
\usepackage{enumitem}
\usepackage{balance}
\usepackage[font={footnotesize}]{caption}
\usepackage{subcaption}
\usepackage{array}
\usepackage{textcomp}
\usepackage{mathtools, nccmath}
\usepackage{graphicx}
\usepackage{amsfonts}
\usepackage{amsmath}
\usepackage{amssymb}
\usepackage{algorithm}
\usepackage{algorithmicx}
\usepackage{tikz}
\usepackage{arydshln}
\usepackage{multirow}
\usepackage{bm}
\usepackage{epstopdf}
\usepackage{siunitx}
\usepackage{xcolor}
\usepackage{tabularx}
\usepackage{cite}
\usepackage{listings}
\usepackage{algpseudocode}
\usepackage{bbold}
\usepackage{hyperref}

\usepackage{booktabs}       
\makeatletter
\let\NAT@parse\undefined
\makeatother
\title{\LARGE \bf General-purpose Clothes Manipulation with Semantic Keypoints}

\author{
Yuhong~Deng$^{1}$,
David~Hsu$^{1,2}$
\thanks{$^{1}$ School of Computing, National University of Singapore, Singapore.\tt\small\{yuhongdeng, dyhsu\}@comp.nus.edu.sg.
}
\thanks{$^{2}$ Smart Systems Institute, National University of Singapore, Singapore.}
}

\begin{document}

\maketitle
\begin{abstract}
Clothes manipulation is a critical capability for household robots; yet,  existing methods are often confined to specific tasks, such as folding or flattening, due to the complex high-dimensional geometry of deformable fabric. This paper presents \textit{CLothes mAnipulation with Semantic keyPoints} (CLASP) for \textit{general-purpose} clothes manipulation, which enables the robot to perform diverse manipulation tasks over different types of clothes.  The key idea of CLASP is semantic keypoints---e.g., ``right shoulder'', ``left sleeve'', etc.---a sparse spatial-semantic representation that is salient for both perception and action. Semantic keypoints of clothes can be effectively extracted from depth images and are sufficient to represent a broad range of clothes manipulation policies. CLASP leverages semantic keypoints to bridge LLM-powered task planning and low-level action execution in a two-level hierarchy. Extensive simulation experiments show that CLASP outperforms baseline methods across diverse clothes types in both seen and unseen tasks. Further, experiments with a Kinova dual-arm system on four distinct tasks---folding, flattening, hanging, and placing---confirm CLASP's performance on a real robot.

\end{abstract}
\IEEEpeerreviewmaketitle

\section{Introduction}
\label{sec:introduction}
\begin{figure}[t]
  \vspace{0.2cm}
  \centering
  \includegraphics[width=\linewidth]{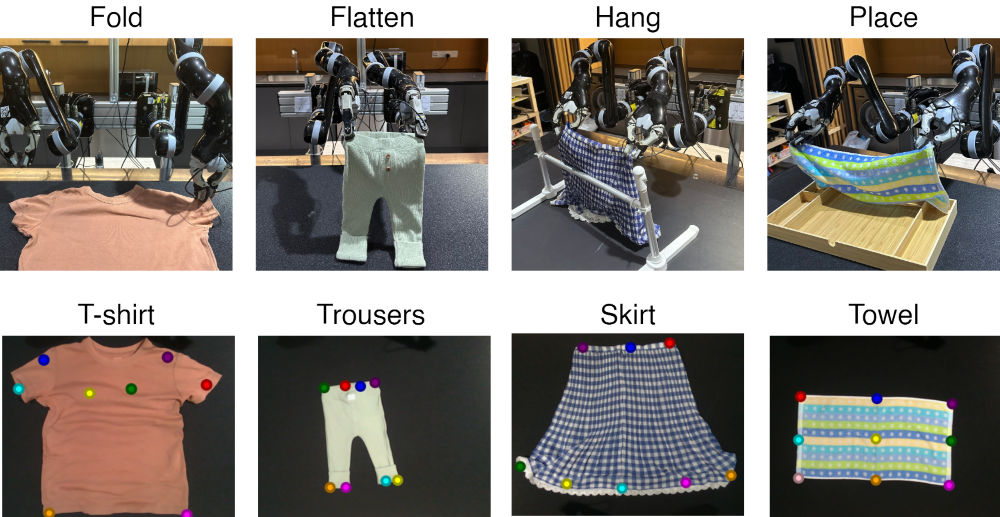}
  \caption{General-purpose clothes manipulation. CLASP performs various manipulation tasks over different types of clothes.}
  \label{fig:task}
  \vspace{-0.7cm}
\end{figure}

Imagine an intelligent household robot taking care of our laundry chores. Toward this goal, we require a general-purpose robot to perform a broad range of clothes manipulation tasks, such as \textit{``fold the T-shirt for storage''} and \textit{``hang the skirt on the hanger''} (Fig.~\ref{fig:task}). Despite recent advances in learning clothes manipulation skills like folding and flattening~\cite{fold_1, flatten_1}, these methods are limited to specific clothes and tasks. Clothes have high-dimensional state~\cite{survey_deformable} and diverse geometric structures, which vary significantly across different categories. The complexity of clothes state and geometries makes it challenging to develop a general-purpose method for manipulating a wide variety of clothes in many different ways.\par

How can we find a state representation for general-purpose clothes manipulation? Considering that clothes have manually designed structures with significant geometric features like sleeves and shoulders, we adopt semantic keypoints on these features as the general spatial-semantic representation. Each semantic keypoint is represented by a language descriptor and its corresponding position. These keypoints, carrying semantic meaning, are relatively easy to extract from observations. Besides, semantic keypoints can (i) define where clothes can be manipulated, aiding high-level task planning, and (ii) capture the clothes' geometry, guiding low-level action execution.\par

To detect semantic keypoints under clothes' self-occlusion and deformation, we use a masked auto-encoder~\cite{mae_st} to learn a spatiotemporal representation through reconstructing masked image sequences. Reconstruction requires the model to infer the geometric structure of clothes from partial observation. To utilize semantic keypoints in clothes manipulation, we develop a general-purpose CLothes mAnipulation method with Semantic keyPoints (CLASP). CLASP is a hierarchical learning method that integrates LLM-powered high-level task planning with low-level action execution based on semantic keypoints.

For high-level task planning, we specify the goals for clothes manipulation using natural language instructions and then employ a large language model (LLM) for task planning. LLM provides commonsense knowledge for task planning, enhancing generalization~\cite{llm_planning}. Specifically, we prompt the LLM to decompose the given language instruction into sequential sub-tasks, where each sub-task is described by an action primitive and a contact point description. The contact point descriptions are selected from the language descriptors of semantic keypoints detected from the observation. Semantic keypoints can define where clothes can be manipulated in task planning. For low-level action execution, we establish a low-level action primitives library consisting of several heuristic policies parameterized by contact point positions. CLASP visually grounds contact points by retrieving from semantic keypoints based on their language descriptors. Based on the LLM proposed action primitive and contact point positions, we invoke a policy from a low-level action primitives library to generate the corresponding trajectory.  Semantic keypoints serve as waypoints to guide trajectory generation. \par 

To evaluate CLASP, we extend the SoftGym benchmark~\cite{softgym} to include 30 tasks across 4 common clothes categories and conduct simulation experiments. CLASP outperforms baseline methods in both seen and unseen tasks. Real-world experiments show that CLASP can be easily transferred to the real world, performing well across a wide variety of clothes and tasks. These results demonstrate that semantic keypoints provide effective cues for clothes manipulation, enabling general-purpose clothes manipulation using our hierarchical learning method with LLM.\par
\section{Related Work}
\label{sec:relatedwork}

\subsection{Deformable Object Manipulation}
There are two main methods for deformable object manipulation: model-based and data-driven methods. Particularly, model-based methods rely on a dynamic model to predict the configurations of deformable objects under a certain action and then select the appropriate action accordingly~\cite{model-base1,model-base2}. However, deformable objects have complex non-linear dynamics, which are challenging to model. Recently, particle-based representation has become a unified solution for deformable object dynamics modeling. However, constructing particles from occluded deformable objects and accurate dynamics modeling is still challenging~\cite{particle-general-1,particle-general-2}.\par
Data-driven methods aim to learn robot actions directly from expert demonstrations without establishing a dynamic model. However, most data-driven methods are designed for specific tasks like rope rearrangement~\cite{rope_1}, cloth folding~\cite{fold_2}, cloth flattening~\cite{flatten_2}, and bag opening~\cite{bag_1}. Toward general-purpose deformable object manipulation, some goal-conditioned methods use goal images to specify different tasks for multi-task learning~\cite{goal_conditioned_1, goal_conditioned_2}. However, goal-conditioned methods often struggle with generalization to unseen tasks. In this paper, we use semantic keypoints to represent the clothes state and propose a hierarchical learning method based on LLM, which performs effectively across a wide range of clothes categories and tasks, while also generalizing to unseen tasks.

\subsection{Language-conditioned Object Manipulation}
 Language provides an intuitive interface in human-robot interaction and can explicitly capture the generalizable concepts between different manipulation tasks. Thus, language-conditioned object manipulation has been widely investigated. Early work focuses on making the robot understand language instructions and perform manipulation tasks~\cite{languge_conditioned_1,languge_conditioned_2}. Recently, foundation models, such as large language models (LLMs) and vision-language models (VLMs), have been utilized in language-conditioned manipulation tasks. These models provide commonsense knowledge for reasoning~\cite{llm_manipulation_1,llm_manipulation_2} and perception~\cite{llm_manipulation_3,rekep}, significantly enhancing generalization capabilities. Most of the previous language-conditioned object manipulation methods focus on rigid objects. However, a significant gap remains between task planning and action execution when it comes to deformable object manipulation. In this paper, we present semantic keypoints as an interface between LLM-powered task planning and low-level action execution, enabling general-purpose clothes manipulation.\par
 
\subsection{State Representation of Deformable Object} 
Given the high-dimensional state of deformable objects, an effective state representation method is necessary. To simulate deformable objects, particles and mesh representations have been explored~\cite{particle_1,particle_2,particle_3}. Such representation provides a solution to model deformable dynamics. However, they tend to be overly dense, making accurate state estimation challenging.
Compared with particles and meshes, keypoints representation has lower dimensionality, leading to more effective policy learning~\cite{keypoint_policy}. Previous keypoints representation focuses on geometric shape, using keypoints to represent the topology of deformable object~\cite{landmark,skeleton}. In contrast, our semantic keypoints are represented by language descriptors and corresponding positions, effectively capturing both semantic and geometric information for manipulation tasks. As a result, our semantic keypoints can bridge and enhance both high-level task planning and low-level action execution.

\begin{figure*}[t]
    \centering
    \vspace{0.3cm}
    \includegraphics[width=\linewidth]{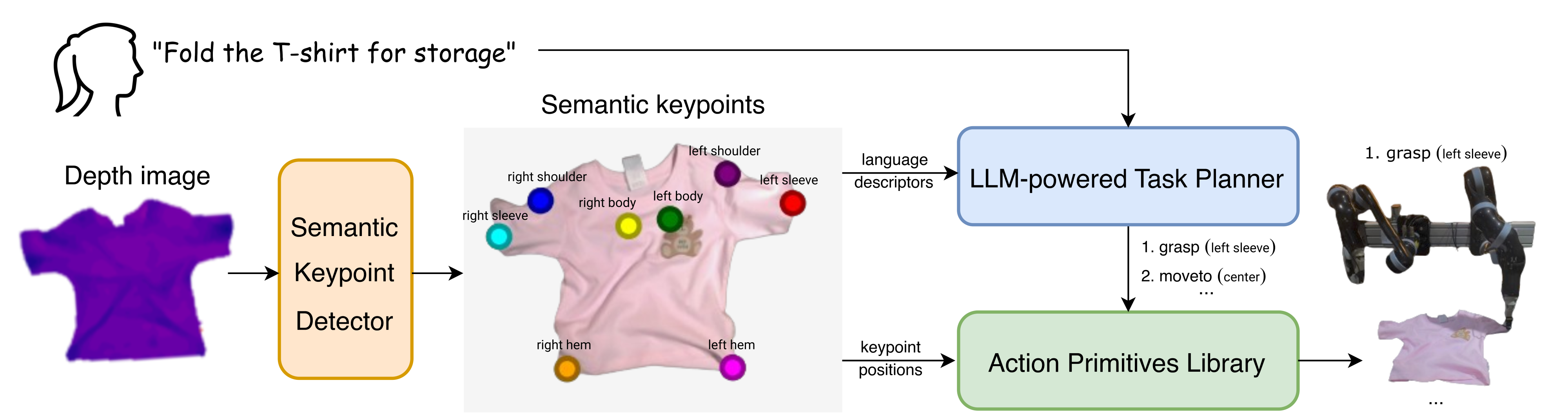}
    \caption{An overview of CLASP.  Given the natural-language task instruction and a depth image, CLASP first detects semantic keypoints, each consisting of a language descriptor and a 2-D geometric position. The task instruction and the language descriptors are fed into an LLM to generate a sequence of sub-tasks on the keypoints. For each sub-task, a low-level action primitives library generates the action on the keypoint position. }
    \label{fig:method}
\end{figure*}

\section{Method}
\label{sec:overview}
In this paper, we propose a general-purpose CLothes mAnipulation method with Semantic keyPoints (CLASP). The key idea of CLASP is using semantic keypoints as a general spatial-semantic representation of clothes. Each semantic keypoint is represented by a language descriptor like \textit{``left sleeve''} and the corresponding keypoint position. The language descriptor can provide semantic information for task planning, while the keypoint position can provide geometry information to guide low-level action execution. \par

Fig.~\ref{fig:method} illustrates the overall framework of CLASP. Given a depth image, a semantic keypoint detector will predict keypoint positions and corresponding language descriptors. The language descriptors of keypoints and the language instruction are fed into an LLM for task planning – inferring a sequence of sub-tasks, where each sub-task is described as an action primitive and a contact point description like \texttt{grasp}(\textit{``left sleeve"}). For each sub-task, CLASP visually grounds contact points to pixel positions by retrieving corresponding semantic keypoints according to the contact point description. Finally, the contact point positions and action primitive will guide low-level action execution based on an action primitives library.\par

In the following sections, we will present how we detect effective semantic keypoints in Sec.~\ref{subsec:keypoint_detection}. Following that, we will introduce how we utilize semantic keypoints in high-level task planning (Sec.~\ref{subsec:task_planning}) and low-level action execution (Sec.~\ref{subsec:action_generation}) for general-purpose clothes manipulation.

\subsection{Semantic Keypoint Detection}
\label{subsec:keypoint_detection}
Although previous work has explored leveraging keypoints for object manipulation~\cite{rekep}, robustly detecting semantic keypoints on clothes remains challenging, especially under self-occlusion and deformation. To address this, we utilize a mask autoencoder (MAE)~\cite{mae_st} as a spatiotemporal learner to establish a powerful latent space. The core idea is that masking acts as a form of occlusion, and recovering the masked areas requires the model to infer geometric structures of clothes from partial observations. Specifically, we first establish a dataset $D = \{\zeta_1, \zeta_2, . . . , \zeta_n\}$ of $n$ sequences with associated discrete-time depth image, keypoints positions, and language descriptors pairs $ \zeta_i = \{(I_t, P_t, S_t)\}_{t=1}^{T}$, where $I_t$ refers to the depth image, $P_t$ and ${S}_t$ refer to the corresponding keypoints positions and language descriptors. We collect such a dataset in simulation through data argumentation on clothes' shape, pose, and configuration.\par
The training process of semantic keypoint detection consists of two stages: the reconstruction stage and the keypoint detection stage. In the first stage, we apply a random masking strategy on the depth image sequence, resulting in masked depth images $\{\tilde{I}_t\}_{t=1}^{T}$. The masked depth images will pass through an encoder-decoder structure to reconstruct the original depth image sequence. Given the reconstructed image sequence $\{\hat{I_t}\}_{t=1}^{T} = f_{dec}(f_{enc}(\{\tilde{I}_t\}_{t=1}^{T}))$, we optimize the
encoder $f_{enc}$ and decoder $f_{dec}$ by minimizing the reconstruction loss, $L_{r} = \sum_{t=1}^{T}\|\hat{I}_t-{I}_t\|$. In the second stage, we fine-tune the encoder $f_{enc}$ with an additional keypoint decoder $f_{kp}$ to predict the semantic keypoints heatmap $\hat{\mathcal{H}}_t = f_{kp}(f_{enc}(I_t))$. The position of the $k$-th keypoint, $p_t^k$, can be determined by $p_t^k= \operatorname{arg max} \hat{\mathcal{H}}_t^k$, where $\hat{\mathcal{H}}_t^k$ corresponds to the $k$-th channel of the semantic keypoints heatmap. For each clothes category, we predefine a set of language descriptors, denoted as $\mathcal{S}$. Given the clothes category, the language descriptor of the $k$-th keypoint, $s_t^{k}$, is obtained by $s_t^{k} = \mathcal{S}_k$. Each channel of the semantic keypoints heatmap  $\hat{\mathcal{H}}_t$ is responsible for detecting keypoints with the same language descriptor. Using the ground truth keypoint positions and their corresponding language descriptors, we generate the ground truth semantic keypoints heatmap $\mathcal{H}_t$. During the second stage, the encoder $f_{enc}$ and keypoint decoder $f_{kp}$ are optimized by minimizing the semantic keypoints heatmap prediction loss, $L_{kp} = \|\hat{\mathcal{H}}_t-\mathcal{H}_t\|$.

\subsection{Task Planning}
\label{subsec:task_planning}
For general-purpose clothes manipulation, we utilize an LLM for task planning due to its powerful commonsense knowledge from extensive internet-scale data. We first use the LLM with a chain of thought prompting~\cite{cot} to define action primitives for clothes manipulation. The LLM is prompted to (i) provide examples of clothes manipulation tasks, (ii) decompose these examples into basic actions, and (iii) summarize the actions used in step (ii) to identify action primitives. In this way, we identify the action primitives set, including \texttt{grasp}, \texttt{moveto}, \texttt{press}, \texttt{release},  \texttt{rotate},  and \texttt{pull}. These action primitives reflect LLM's commonsense knowledge, enhancing task planning. To generate sub-tasks, we prompt the LLM with examples of language instructions paired with desirable sub-tasks sequences. Given a language instruction and language descriptors of semantic keypoints, the LLM generates sub-tasks by selecting action primitives from predefined action primitives set and contact points from language descriptors of semantic keypoints.

\subsection{Action Execution}
\label{subsec:action_generation}
To complete each sub-task, we establish a low-level action primitives library.  In the low-level action primitives library, there are some heuristic rules to create waypoints based on the contact point positions and selected action primitives. Once the waypoints are determined, a motion planning algorithm is applied to generate the complete trajectory for the robot’s execution. Specifically, the action primitive \texttt{grasp} picks up clothes parts through planar grasping with one or both arms. The action primitive \texttt{moveto} transports the clothes to a target position, and target positions are determined by semantic keypoints or the location of the receptacle (e.g., a hanger or a box), depending on the task context. The action primitive \texttt{press} smooths out wrinkles
by pressing along the normal vector of the table. The pressing distance is fixed at 1 cm in this paper. The action primitive \texttt{release} drops clothes parts by opening grippers. The action primitive \texttt{rotate} rotates the clothes grasped by grippers. This primitive is frequently used to align the clothes with the hanger in hanging tasks, and the hanger's pose can be obtained using an open-vocabulary object detector. The action primitive \texttt{pull} stretches the clothes to flatten them. In this paper, we set the stretch distance to 10\% of the distance between the grippers of two arms.
\section{Experiments}
\label{sec:experiments}
We aim to answer the following research questions: 
(i) How well does our semantic keypoint detector detect
effective semantic keypoints? (Sec.~\ref{subsec:keypoint_expri}) (ii) Can CLASP perform well and generalize to a wide variety of clothes and manipulation tasks? (Sec.~\ref{subsec:simu_expri}) (iii) Can CLASP trained in simulation be transferred to real-world clothes manipulation tasks? (Sec.~\ref{subsec:real_expri})

\subsection{Semantic Keypoint Detection Experiments}
\label{subsec:keypoint_expri}

\begin{table}
\centering
\caption{ Semantic keypoint detection performance.}
\resizebox{0.95\linewidth}{!}{
\begin{tabular}[t]{lcccc}
\toprule
&AKD (pixel) $\downarrow$ &$\textit{AP}_8$ (\%) $\uparrow$ 
&$\textit{AP}_4$ (\%) $\uparrow$ &$\textit{AP}_2$ (\%) $\uparrow$
\\
\midrule
DINOv2-S & 16.3 &63.3 &39.7 &15.6 \\
DINOv2-B & 12.6 &68.8 &42.5 &16.9\\
DINOv2-L &12.1 &70.2 &42.5 &17.0\\
DINOv2-g &10.0 &75.2 &47.5 &18.7 \\
MAE (from scratch) &5.3 &84.4&66.2&41.1\\
Ours &\textbf{3.8}&\textbf{91.0}&\textbf{75.4}&\textbf{50.2}\\
\bottomrule
\end{tabular}}
\label{tab:keypoint}
\end{table}%

To evaluate the performance of our semantic keypoint detector, we compare it with several baseline methods.  Baseline methods include DINOv2~\cite{dinov2} and MAE (from scratch). DINOv2 is a large pretrained vision model with powerful visual features. We directly use the features from DINOv2 and train a keypoint decoder to predict semantic keypoints. We evaluate four types of DINOv2 with increasing parameters: DINOv2-S (Small), DINOv2-B (Base), DINOv2-L (Large), and DINOv2-g (Giant). Larger models offer better performance but come with increased computational burden. MAE (from scratch) uses the same architecture as our method but is trained from scratch without a reconstruction stage. All baseline methods are trained for 200 epochs. Our model is trained for 100 epochs in both the reconstruction and keypoint detection stages. We use two standard metrics for keypoint detection: Average Keypoint Distance (AKD) and Average Precision (AP). AKD measures the average distance between ground truth and detected keypoints, while AP represents the proportion of correctly detected keypoints under the given threshold. Given that the observed depth images have a resolution of 224 × 224, we set thresholds at 8, 4, and 2 pixels.\par

\begin{figure}[t]
    \vspace{0.3cm}
    \centering
    \includegraphics[width=\linewidth]{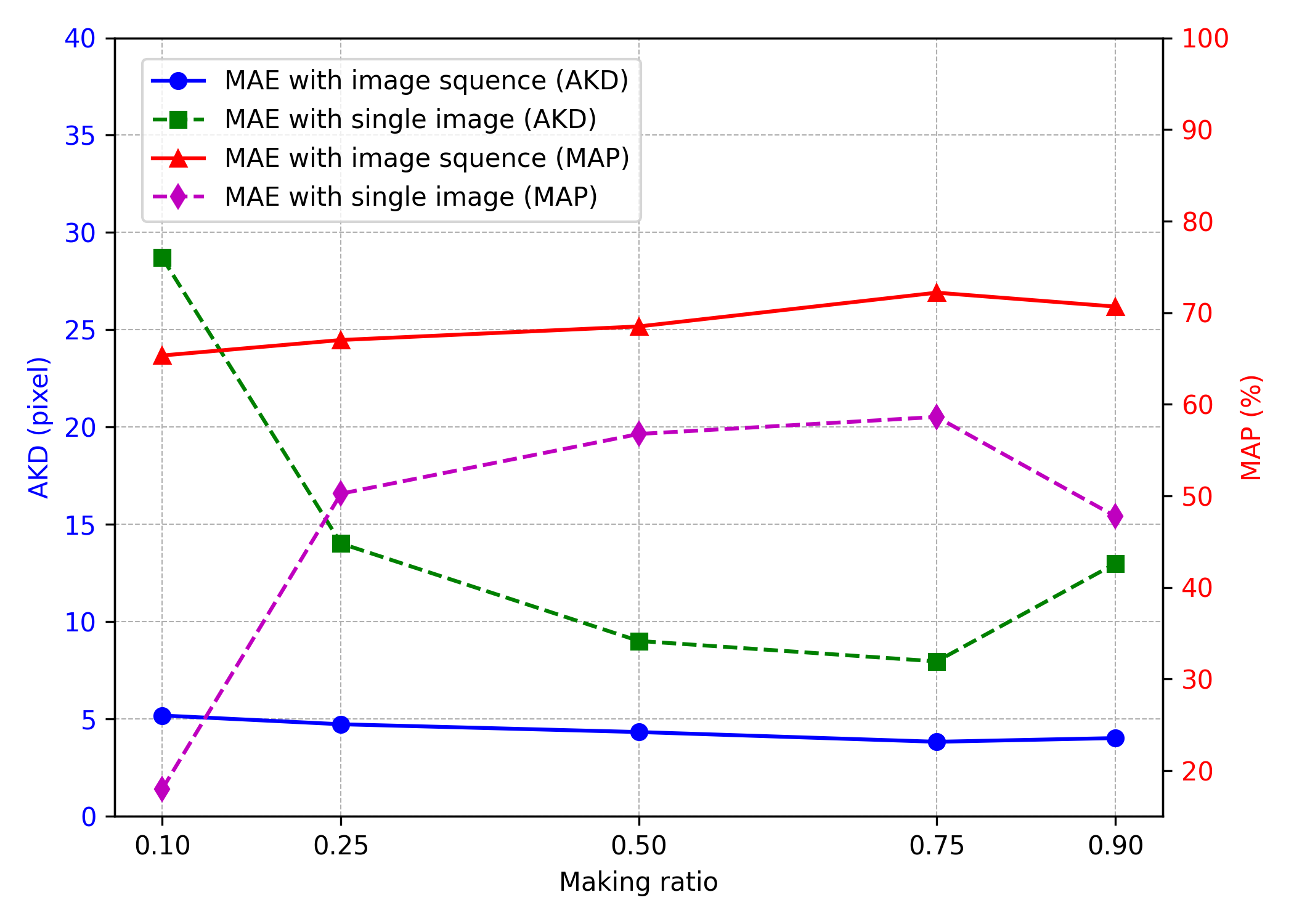}
    \caption{An ablation study on the effects of masking and temporal information on semantic keypoint detection.}
    \label{fig:keypoint}
\end{figure}

The results are shown in TABLE~\ref{tab:keypoint}. Our method outperforms both baseline methods, demonstrating that the reconstruction stage effectively learns a powerful representation, which enhances the performance of semantic keypoint detection. To further analyze our method, we conduct an ablation study by varying the masking ratio and comparing training with image sequences versus a single image. The evaluation metrics include Average Keypoint Distance (AKD) and Mean Average Precision (MAP), where MAP represents the mean of average precision values calculated at different thresholds. As Fig.~\ref{fig:keypoint} shows, increasing the masking ratio allows the model to better infer the geometric structure of clothes under self-occlusion, leading to performance improvement on keypoint detection. The model achieves the best performance when the masking ratio reaches 0.75, beyond which the information becomes insufficient. Additionally, reconstructing image sequences yields better results than reconstructing a single image, as the temporal information helps the model capture consistent geometric features relevant to semantic keypoints. In summary, our model design enables the establishment of a powerful spatiotemporal representation, significantly improving the model's ability to detect semantic keypoints of clothes.

\subsection{Simulation Experiments}
\label{subsec:simu_expri}

\begin{table*}[t]
\vspace{0.2cm}
\centering
\small
\caption{Simulation experiments. The average success rates (\%) of CLASP and baseline methods on clothes manipulation. }
\resizebox{0.95\linewidth}{!}{
\begin{tabular}{@{}lc cc cc cc cc ccc@{}}
\toprule
\multirow{2}*{Method} 
& \multicolumn{2}{c}{\begin{tabular}[c]{@{}c@{}}  Flattening
\\(unseen object) \end{tabular}} 
& \multicolumn{2}{c}{\begin{tabular}[c]{@{}c@{}}  Hanging
\\ (unseen object) \end{tabular}} 
& \multicolumn{2}{c}{\begin{tabular}[c]{@{}c@{}}  Placing
\\ (unseen object) \end{tabular}} 
& \multicolumn{2}{c}{\begin{tabular}[c]{@{}c@{}}  Folding
\\ (unseen object) \end{tabular}}
& \multicolumn{3}{c}{\begin{tabular}[c]{@{}c@{}}  Folding
\\ (unseen requirement) \end{tabular}} 
\\
\cmidrule(lr){2-3} \cmidrule(lr){4-5} \cmidrule(lr){6-7} 
\cmidrule(lr){8-9} \cmidrule(lr){10-12}
& Trousers &Towel &T-shirt &Skirt &Trousers &T-shirt &Trousers
&Skirt &Position &Direction &Times \\
\midrule
CLIPORT~\cite{languge_conditioned_2} &8.3 &9.2 &76.7 &66.7 &70.0 &73.3 &0.0 &0.0 &76.7 &65.0 &0.0\\
Goal-conditioned Transporter~\cite{goal_conditioned_2} &10.0 &6.7  &36.7 &40.0
 &36.7 &60.0  &0.0 &0.0 &8.3 &0.0 &0.0 \\
FlingBot~\cite{flingbot} &29.2 &34.2 & N/A  &N/A &N/A &N/A &N/A &N/A
&N/A &N/A &N/A\\
FabricFlowNet~\cite{fabricflownet}  &N/A &N/A &N/A &N/A &N/A &N/A
&0.0 &2.5 &30.0 &3.3 &0.0\\
CLASP &\textbf{50.0} &\textbf{55.0} &  \textbf{93.3} & \textbf{73.3} &\textbf{93.3}& \textbf{93.3} &\textbf{86.7} &\textbf{83.3} &\textbf{100.0} &\textbf{96.7} &\textbf{95.0}\\

\midrule
\multirow{2}*{Method} 
& \multicolumn{2}{c}{\begin{tabular}[c]{@{}c@{}}  Flattening
\\(seen object) \end{tabular}} 
& \multicolumn{2}{c}{\begin{tabular}[c]{@{}c@{}}  Hanging
\\ (seen object) \end{tabular}} 
& \multicolumn{2}{c}{\begin{tabular}[c]{@{}c@{}}  Placing
\\ (seen object) \end{tabular}} 
& \multicolumn{2}{c}{\begin{tabular}[c]{@{}c@{}}  Folding
\\ (seen object) \end{tabular}}  \\
\cmidrule(lr){2-3} \cmidrule(lr){4-5} \cmidrule(lr){6-7} 
\cmidrule(lr){8-9} 
&T-shirt &Skirt &Trousers &Towel
&Towel  &Skirt &Towel &T-shirt

 \\
\midrule
CLIPORT~\cite{languge_conditioned_2} &32.5 &36.7 &76.7 &83.3 &93.3 &\textbf{93.3} &77.5 &80.0 \\
Goal-conditioned Transporter~\cite{goal_conditioned_2} &26.7& 33.3
& \textbf{100.0} & 80.0 &66.7 &83.3 &83.3 &76.7  \\
FlingBot~\cite{flingbot} &\textbf{66.7} &\textbf{85.0} &N/A &N/A &N/A &N/A &N/A &N/A   \\
FabricFlowNet~\cite{fabricflownet}  &N/A &N/A &N/A &N/A &N/A &N/A
 &93.7 & \textbf{100.0} \\
CLASP  & 55.0 & 78.3 & 96.7 & \textbf{96.7} & \textbf{96.7} &
90.0 &\textbf{100.0} &\textbf{100.0} \\

\bottomrule
\end{tabular}}
\label{tab:exp_resu}
\vspace{-0.25cm}
\end{table*}
To evaluate the proposed method's performance on clothes manipulation tasks, we conduct experiments in SoftGym~\cite{softgym}, where clothes are modeled as particles with ground truth positions. 3D models of clothes are sampled from CLOTH3D~\cite{cloth3d} dataset, covering 4 common clothes categories in human's daily life: T-shirts, trousers, skirts, and towels. For each category, over 35 instances of varying sizes and shapes are used. In addition, we extend the SoftGym benchmark to 30 tasks. These tasks can be divided into 4 categories:\par

\begin{itemize}
    \item \textbf{Folding} tasks involve folding clothes to achieve a target configuration. The success of a folding task is determined by the particle position error between the folded item and the target configuration.\par
    \item  \textbf{Flattening} tasks involve flattening crumpled clothes with random deformations. The success of a flattening task is determined by the coverage area of the clothes. \par
    \item \textbf{Hanging} tasks require hanging clothes on a hanger. A hanging task is successful when the clothes are fully hung on the hanger without any part touching the ground. \par
    \item \textbf{Placing} tasks require placing the clothes in a box. A placing task is successful when the clothes are laid flat inside the box.\par  
\end{itemize}

In our experimental setup, only half of tasks are seen during training through examples in the prompt or demonstrations. Unseen tasks involve new object categories and new requirements like folding direction, position, and times. For each task category and object category, we conduct 120 trials with different clothes configurations to calculate the success rate. Baseline methods include two general multi-task learning frameworks (CLIPORT and Goal-conditioned Transporter) and two task-specific algorithms (FlingBot and FabricFlowNet): \par

\begin{itemize}
    \item \textbf{CLIPORT}~\cite{languge_conditioned_2} represents a typical end-to-end algorithm for learning language-conditioned manipulation policies, which leverages a pretrained vision-language model.
    \item \textbf{Goal-conditioned Transporter}~\cite{goal_conditioned_2} represents a typical goal-conditioned transporter network for deformable object manipulation, which infers the optimal action from the current and goal images.
    \item \textbf{FlingBot}~\cite{flingbot} is a self-supervised learning framework designed for flattening crumpled clothes using a fling action.
    \item \textbf{FabricFlowNet}~\cite{fabricflownet} is a learning framework designed for clothes folding, which infers actions based on optical flow.
\end{itemize}

The experiment results are shown in TABLE~\ref{tab:exp_resu}. Overall, our method outperforms the two multi-task learning methods on both seen and unseen tasks. Compared with Goal-conditioned Transporter, CLIPORT shows better generalization. Language instructions facilitate capturing similarities between different tasks and the pretrained vision-language model enables CLIPORT to capture such similarities. However, CLIPORT's generalization is limited when adapting to tasks with significantly different action sequences, such as transferring the skill from folding a T-shirt to folding trousers, because it learns task-specific manipulation policies in an end-to-end manner. In contrast, CLASP is a hierarchical learning framework that learns generalizable language and visual concepts across a wide range of clothes manipulation tasks. The commonsense knowledge from LLM allows CLASP to handle unseen tasks by decomposing them into predefined action primitives. Furthermore, semantic keypoints are task-agnostic, providing cues for task planning and action execution in unseen manipulation tasks.\par
Compared to the two task-specific algorithms, CLASP shows comparable performance on seen tasks, demonstrating the effectiveness of the proposed method for clothes manipulation. On unseen tasks, task-specific algorithms tend to fail due to distribution shifts. In contrast, CLASP performs well on unseen tasks.

\subsection{Real-Robot Experiments}
\label{subsec:real_expri}

\begin{figure}
    \centering
    \includegraphics[width=\linewidth]{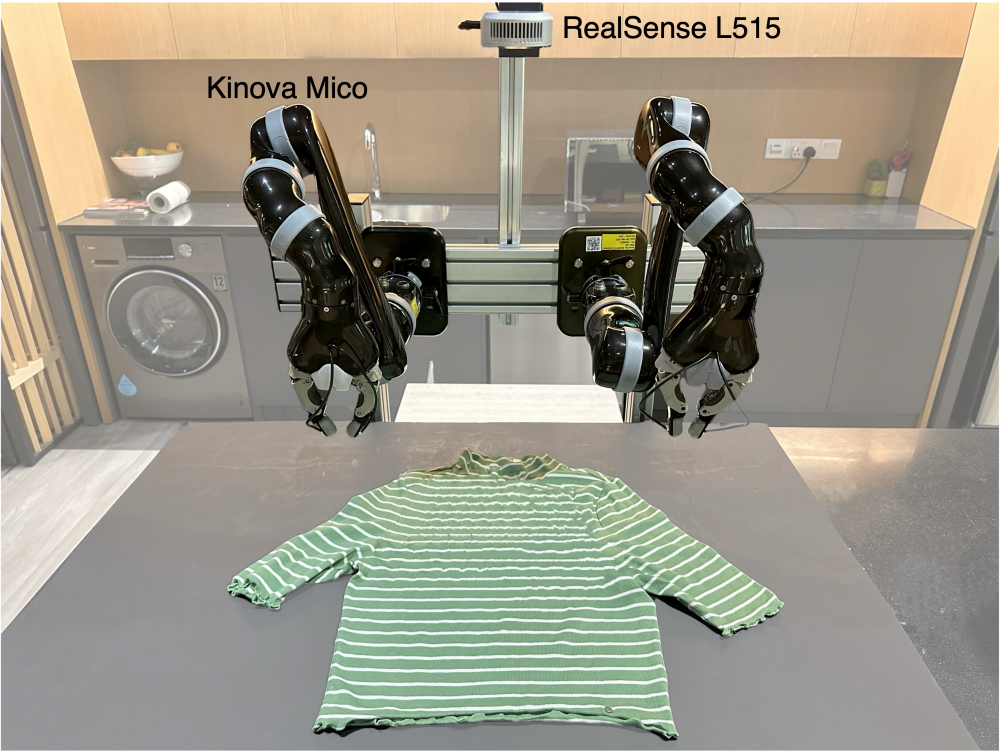}
    \caption{Setup for real-robot experiments. The dual-arm system consists of an Intel RealSense camera for depth sensing and two Kinova Mico arms.}
    \label{fig:system}
    \vspace{-0.3cm}
\end{figure}

We utilize depth images instead of RGB images to ensure our method can be directly transferred to real-world scenarios. To evaluate the sim-to-real performance of our method, we establish a dual-arm robot manipulation system. As Fig.~\ref{fig:system} shows, the system consists of two Kinova Mico robot arms and a top-down RealSense L515 RGB-D camera for capturing depth images. The clothes are placed on a platform in front of the robot. To generate the dual-arm trajectories, we utilize a motion planning algorithm using MoveIt!~\cite{moveit} to avoid collisions and synchronize the motion between two arms. To evaluate the performance of our method, we select a diverse set of clothes varying in size, appearance, and shape, ranging from infant trousers to adult shorts. \par

\begin{figure*}
    \centering
    \vspace{1cm}
    \includegraphics[width=\linewidth]{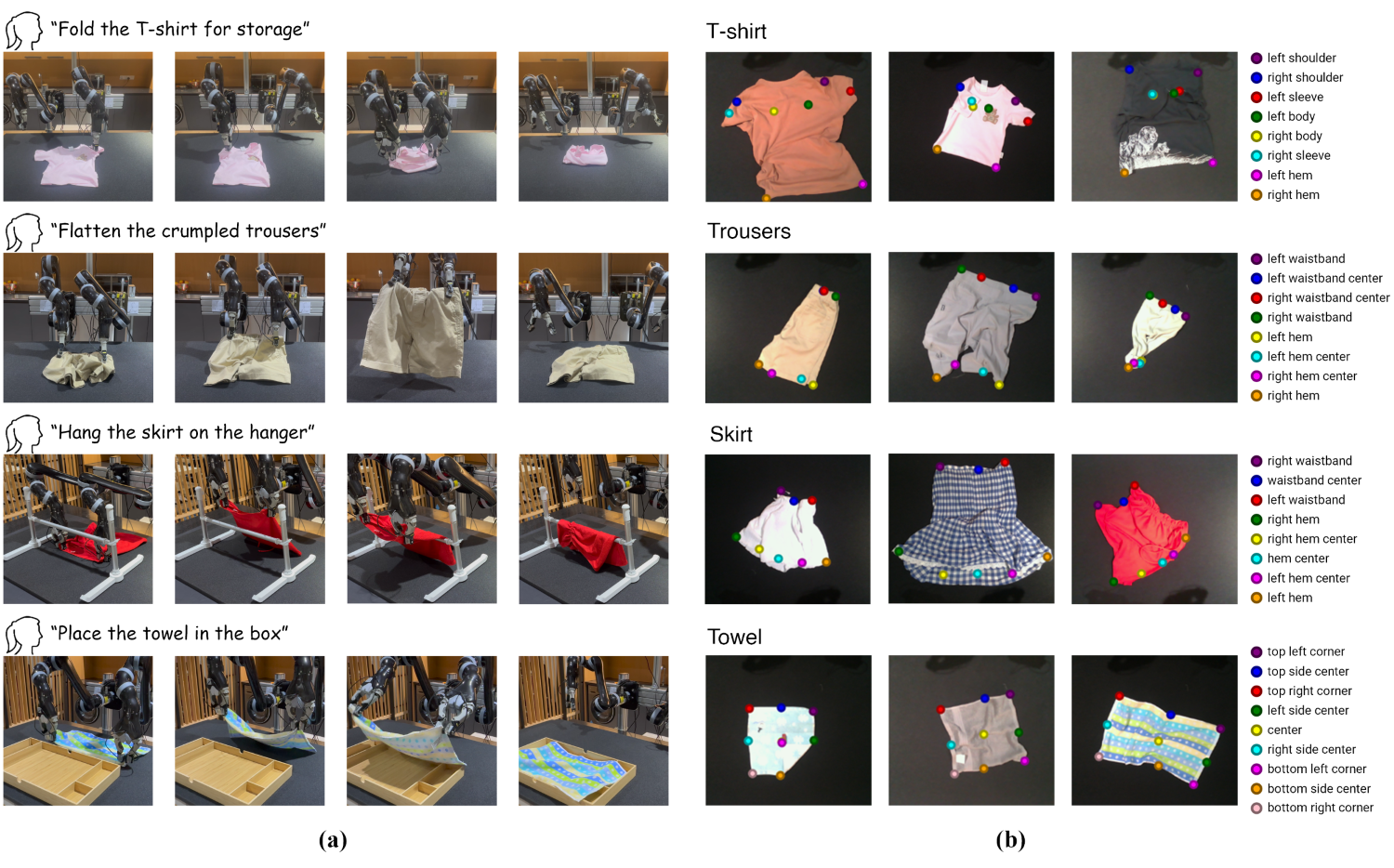}
    \caption{Real-robot experiments. (a) Four clothes manipulation tasks: folding, flattening, hanging, and placing. (b) Semantic keypoint detection on a variety of clothes. }
    \label{fig:experiment}
\end{figure*}

We first evaluate the performance of our semantic keypoint detector, which is trained in simulation. To mitigate noise from depth sensing in real-world environments, we use SAM~\cite{SAM} and OWLv2~\cite{OWLv2} to generate masks that filter out noisy depth data and smooth the images. The experimental results are shown in Fig.~\ref{fig:experiment} (b). Our semantic keypoint detector can robustly detect keypoints across a diverse range of clothes, even under irregular deformation and occlusion, without requiring fine-tuning on real-world images.\par

We further test the performance of CLASP on some real-world clothes manipulation tasks. Fig.~\ref{fig:experiment} (a) illustrates four representative examples:\textit{``Fold the T-shirt for storage''}, \textit{``Flatten the crumpled trousers"}, \textit{``Hang the skirt on the hanger"} and \textit{``Place the towel in the box"}. CLASP demonstrates the ability to manipulate various clothes in many different ways in real-world scenarios. 
\section{Conclusion}
\label{sec:conclusion}
In this paper, we propose semantic keypoints as a general spatial-semantic representation of clothes and enable general-purpose clothes manipulation using our hierarchical learning method with LLM. To detect semantic keypoints, we use a masked autoencoder to establish a spatiotemporal representation capable of handling self-occlusion and deformation. The integration of commonsense knowledge from the LLM and the general semantic keypoint representation ensures the generalization of our method. Simulation experiment results show that our method performs well on both seen and unseen tasks, including new object categories and task requirements. Furthermore, the proposed method can be directly transferred to real-world scenarios and performs well on a wide variety of clothes. However, our current solution is an open-loop system, where the task planning is performed
only once at the beginning of the manipulation task,
making it sensitive to unexpected state changes or external
disturbances. Future improvement will focus on developing a closed-loop pipeline based on our semantic keypoints representation.

\section*{Acknowledgment}
 This research is supported by the National Research Foundation, Singapore, under its Medium Sized Centre Program, Center for Advanced Robotics Technology Innovation (CARTIN).

\newpage  


\end{document}